\pdfoutput=1

\documentclass[11pt]{article}

\usepackage[final]{acl}

\usepackage{tablefootnote}
\usepackage{xcolor}
\usepackage{times}
\usepackage{latexsym}
\usepackage{multirow}
\usepackage{multicol}
\usepackage{booktabs}
\usepackage{pifont}
\usepackage{listings}

\lstdefinestyle{mystyle}{
  backgroundcolor=\color{lightgray},   
  commentstyle=\color{green},
  keywordstyle=\color{blue},
  numberstyle=\tiny\color{gray},
  stringstyle=\color{red},
  basicstyle= \ttfamily\scriptsize,
  breakatwhitespace=false,         
  breaklines=true,                 
  captionpos=b,                    
  keepspaces=true,                 
  numbers=left,                    
  numbersep=4pt,                  
  showspaces=false,                
  showstringspaces=false,
  showtabs=false,                  
  tabsize=2,
  extendedchars=true,
  escapeinside={(*@}{@*)}
}

\usepackage[normalem]{ulem}
\useunder{\uline}{\ul}{}
\usepackage[T1]{fontenc}

\usepackage[utf8]{inputenc}
\usepackage{microtype}

\usepackage{inconsolata}

\usepackage{graphicx}

\title{Bi’an: A Bilingual Benchmark and Model for Hallucination Detection in Retrieval-Augmented Generation}

\author{Zhouyu Jiang\ , Mengshu Sun\ , Zhiqiang Zhang\ , Lei Liang  
\\ 
Ant Group \\
    \texttt{\{jiangzhouyu.jzy,mengshu.sms,lingyao.zzq,leywar.liang\}@antgroup.com}}
\usepackage{CJKutf8}
\begin{document}
\maketitle

\begin{abstract}
Retrieval-Augmented Generation (RAG) effectively reduces hallucinations in Large Language Models (LLMs) but can still produce inconsistent or unsupported content. Although LLM-as-a-Judge is widely used for RAG hallucination detection due to its implementation simplicity, it faces two main challenges: the absence of comprehensive evaluation benchmarks and the lack of domain-optimized judge models. To bridge these gaps, we introduce \textbf{Bi'an}, a novel framework featuring a bilingual benchmark dataset and lightweight judge models. The dataset supports rigorous evaluation across multiple RAG scenarios, while the judge models are fine-tuned from compact open-source LLMs. Extensive experimental evaluations on Bi'anBench show our 14B model outperforms baseline models with over five times larger parameter scales and rivals state-of-the-art closed-source LLMs. We will release our data and models soon at \url{https://github.com/OpenSPG/KAG}.

\end{abstract}

\section{Introduction}
The practical deployment of large language models (LLMs) is significantly hindered by the persistent issue of "hallucination," wherein the model generates outputs that deviate from factual accuracy \citep{dziri-etal-2022-origin, varshney2023stitchtimesavesnine, li-etal-2024-dawn}. To address this limitation, Retrieval-Augmented Generation (RAG) techniques have been developed, which enhance model performance by incorporating external data sources \citep{10.5555/3495724.3496517, Borgeaud2021ImprovingLM, jiang-etal-2024-efficient}. Despite these advancements, RAG systems are not immune to generating outputs that may lack substantiation or contradict the provided references \citep{mallen-etal-2023-trust, 10.1145/3637528.3672065}. In the context of RAG hallucination detection, the LLM-as-a-Judge approach has emerged as a widely adopted solution due to its simplicity \citep{saad-falcon-etal-2024-ares, ravi2024lynxopensourcehallucination, es-etal-2024-ragas}. This method directly leverages an LLM to assess whether the RAG system's output aligns with the input text.

However, the application of LLM-as-a-Judge in RAG hallucination detection encounters two primary limitations that hinder its effectiveness. Firstly, the absence of comprehensive benchmark datasets poses a significant constraint on the rigorous evaluation of model performance. Although several relevant datasets are available \citep{li-etal-2023-halueval, niu-etal-2024-ragtruth, ravi2024lynxopensourcehallucination}, they exhibit notable limitations in linguistic diversity, domain coverage, and task complexity. Secondly, the paucity of specialized training data has resulted in a critical gap in specialized judge model development. This deficiency necessitates reliance on large-scale or advanced closed-source models for practical implementations, consequently leading to substantial operating expenses.

To overcome these issues, we introduce \textbf{Bi'an}, a comprehensive framework comprising a bilingual (Chinese-English) benchmark, Bi'anBench, for RAG hallucination detection and optimized lightweight judge models. For the construction of Bi'anBench, we have selected four types of RAG scenarios suitable for evaluation: question answering, summarization, Data-to-Text, and machine translation. Following the extensive curation of open-source datasets in Chinese and English, we create two synthetic data generation pipelines to produce high-quality test cases, resulting in a large-scale dataset with 22,992 instances. For model development, we implement a stratified sampling approach from source datasets in Bi'anBench and an ensemble-based approach to construct training samples, generating fine-tuning data and preference pairs simultaneously. The final model training involves sequential supervised fine-tuning and Direct Preference Optimization (DPO) \citep{rafailov2023direct}, leveraging the open-source Qwen2.5 architecture (7B and 14B variants) \citep{qwen2.5}.
\begin{figure*}[htbp]
    \centering
    \includegraphics[width=0.97\linewidth]{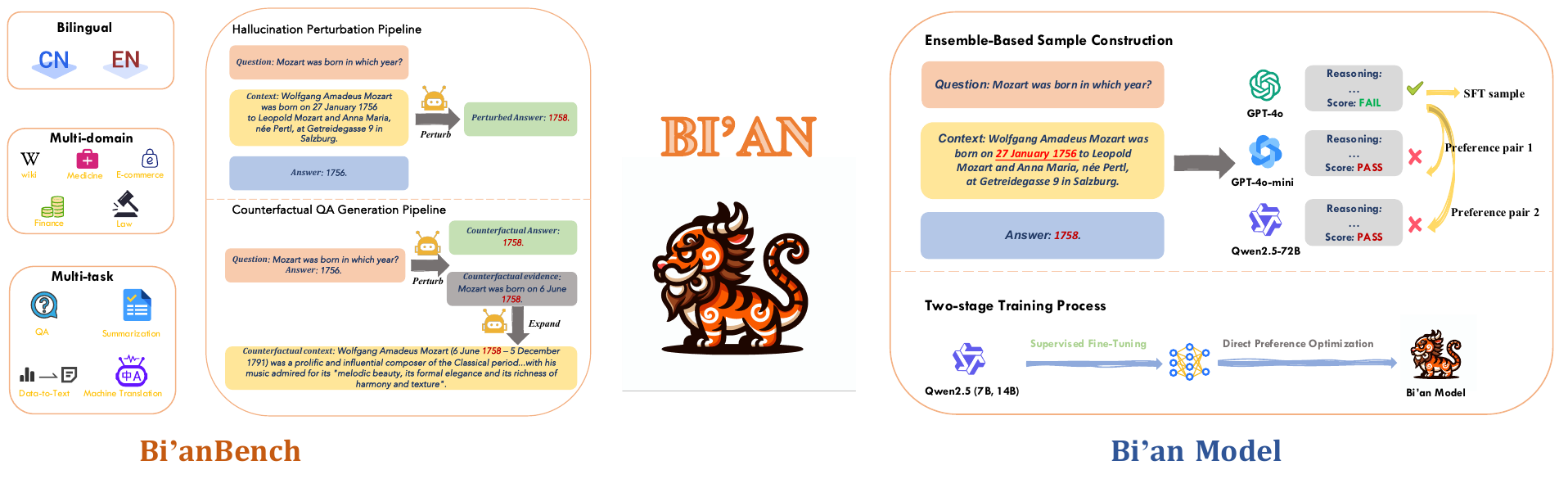}
    \caption{An overview of the Bi'an framework, including Bi'anBench and Bi'an Model. In Chinese mythology, Bi'an is the offspring of a dragon and a tiger, a mythical creature capable of discerning right from wrong, thus aligning with the scenario of RAG hallucination detection.}
    \label{fig:overview}
\end{figure*}

Experimental results on Bi'anBench show that our 14B model outperforms Qwen2.5-72B-Instruct and approaches the capabilities of the state-of-the-art closed-source model GPT-4o. Ablation studies confirm the performance gains from our training methodology, validating the effectiveness of each component.

\section{Methodology}
\subsection{Task Definition}
Referring to the definition from the previous work HaluBench \citep{ravi2024lynxopensourcehallucination}, we define hallucination in RAG as the situation where, given an input query, the output contains information that is either unsupported or contradictory relative to the context. A notable distinction is that, while the definition in HaluBench does not take into account the query information, we have relaxed this constraint. It is worth noting that the definition of hallucination for LLMs is not uniform \citep{10.1145/3703155}; factual judgments can be made relative to real-world facts or the input information. In our work, we focus solely on the latter, \textit{faithfulness hallucination}, as it aligns with the motivation behind RAG systems and avoids the impact of the dynamic nature of real-world facts on evaluation stability.

\subsection{Benchmark Construction}
\label{sec:bench}
\begin{table}[htbp]
\scalebox{0.57}
{
\begin{tabular}{lcccc}
\hline
Benchmark  & \multicolumn{1}{l}{Multilingual} & \multicolumn{1}{l}{Multi-domain} & \multicolumn{1}{l}{Multi-task} & \multicolumn{1}{l}{Size} \\ \hline
HaluEval \citep{li-etal-2023-halueval}  & \ding{56} & \ding{56} & \ding{52} & 30,000 \\
RAGTruth \citep{niu-etal-2024-ragtruth}   & \ding{56}& \ding{56} & \ding{52} & 2,965 \\
HaluBench \citep{ravi2024lynxopensourcehallucination}  & \ding{56} & \ding{52} & \ding{56} & 14,900 \\
\textbf{Bi'anBench} & \ding{52} & \ding{52} & \ding{52} & 22,992 \\ \hline
\end{tabular}
}
\caption{A comparison of Bi'anBench with other RAG hallucination detection datasets.}
\label{tab:comparison}
\end{table}
Here, we introduce the construction process for Bi'anBench. Statistical analysis, prompt templates, and additional technical details are available in the Appendix \ref{bench_detail}. A comparison of Bi’anBench with other relevant benchmarks is presented in Table \ref{tab:comparison}.

\noindent \textbf{Data Collection}
To comprehensively evaluate LLMs' ability to detect RAG hallucinations, we have extended existing benchmarks across three critical dimensions: linguistic diversity, domain coverage, and task complexity. Bi'anBench incorporates bilingual data (Chinese and English) and encompasses four fundamental RAG tasks: question answering, summarization, Data-to-Text, and machine translation. All source datasets are curated from open-source repositories, and the majority have undergone peer review to ensure data reliability. Partially based on HaluBench, we meticulously collect and organize 15 English task subsets alongside 10 Chinese task subsets. Notably, we deliberately exclude creative writing tasks as their subjective nature makes verification against input information impossible, conflicting with our operational definition of RAG hallucination detection.

\noindent \textbf{Hallucination Perturbation Pipeline}
Recognizing the scarcity of specialized data for RAG hallucination detection, we create a hallucination perturbation pipeline to generate test cases following our initial RAG dataset collection. Specifically, we employ GPT-4o as our primary synthesis model, utilizing \texttt{(question,context,answer) } triplets as inputs. The model is guided to introduce semantically plausible yet factually inconsistent modifications to the original answers. The resulting \texttt{(question,context,perturbed\_answer)} triplets labeled as \texttt{`FAIL'} constitute our core test cases for hallucination detection evaluation. We apply this pipeline to approximately half of each relevant data subset.

\noindent \textbf{Counterfactual QA Generation Pipeline}
To enable RAG hallucination detection using context-free QA datasets, we develop a GPT-4o-based counterfactual QA generation pipeline comprising two stages. Initially, the model synthesizes counterfactual answers with brief supporting evidence from existing QA pairs. Subsequently, it expands the evidence into detailed contexts of desired style and length. This process yields both positive instances \texttt{(question,counterfactual context,counterfactual answer)} and negative instances \texttt{(question,counterfactual context,factual answer)} for model evaluation. Given the pipeline's complexity and computational demands, we limit its use to generate a specialized subset for investigating how LLMs' parametric knowledge affects RAG hallucination detection.

Based on the aforementioned methods, we obtain
the Bi'anBench, which includes a total of 22,992 test cases, divided into three subsets: the English subset Bi'anBench\_EN (13,301), the Chinese subset Bi'anBench\_ZH (7,757), and a special counterfactual QA subset Bi'anBench\_CF (1,934).

\subsection{Model Development}
The details of the model development, including the prompt templates used for training set construction and the hyperparameters for model training, etc., are included in Appendix \ref{app:model_dev}.

\noindent \textbf{Training Dataset Construction} We develop a stratified sampling strategy to prepare training data while ensuring out-of-distribution generalization testing. This approach involves categorizing the source datasets of Bi'anBench by task and selectively sampling subsets from each category. The sampled data differs from the existing data in Bi'anBench and is then processed through our hallucination perturbation pipeline to generate structured quadruples \texttt{(question,context,answer,label)} as RAG hallucination detection requests.

Subsequently, we implement an ensemble-based sample construction approach that employs three models (GPT-4o-0806, GPT-4o-mini, and Qwen2.5-72B-Instruct) in parallel with consistent prompt templates. Each model handles RAG hallucination detection requests independently, generating JSON outputs containing reasoning processes and judgment results. These judgment results are then compared against true labels to create 1) SFT samples from correct predictions and 2) preference pairs from conflicting results. 


Two key implementation rules apply: a) When multiple correct outputs exist, selection follows the priority order: GPT-4o > Qwen2.5-72B-Instruct > GPT-4o-mini. b) Samples with unanimous model errors are excluded from training data.

Ultimately, we construct 5,994 SFT samples and 1,713 preference pairs for model training.

\noindent \textbf{Two-stage Training Process} Following the general paradigm of post-training for LLMs \citep{chatgpt2022, qwen2.5, llama3modelcard}, we also adopt a two-stage training process for the Bi'an model, which includes supervised fine-tuning and preference learning based on Direct Preference Optimization (DPO) \citep{rafailov2023direct}. Due to computational resource constraints, we utilize the popular parameter-efficient fine-tuning method, LoRA \citep{hu2022lora}, for our training. After completing the SFT, we merge the obtained LoRA adapter with the original model before proceeding to DPO training. We conduct distributed training using 4 Nvidia A100-80G GPUs for two lightweight versions of Qwen2.5 (7B and 14B), facilitated by DeepSpeed \citep{10.1145/3394486.3406703}.
\section{Experiments}
\subsection{Experimental Settings}
\textbf{Datasets}: We conduct experiments on two core subsets of Bi'anBench: Bi'anBench\_EN and Bi'anBench\_ZH. 

\noindent \textbf{Baselines:} We test a range of strong models for comparison, including the closed-source GPT-4o series and the open-source Llama and Qwen family. Additionally, we test Lynx-8B-v1.1 \citep{ravi2024lynxopensourcehallucination} on Bi'anBench\_EN, which is an open-source model specifically fine-tuned for hallucination detection in English QA, based on Llama3. 

\noindent \textbf{Evaluation Metric:} We employ accuracy as the metric, parsing the output JSON strings to extract the judgment results and matching them with the true labels to determine accuracy.

\subsection{Main Results}

\begin{table*}[]
\centering
\small
\scalebox{0.78}{
\begin{tabular}{cccccccccc}
\toprule
Model                 & \multicolumn{4}{c}{Bi'anBench\_EN}                                  &               & \multicolumn{3}{c}{Bi'anBench\_ZH}                                      &               \\ \cline{2-5} \cline{7-9}
                      & QA            & Summarization & Data-to-Text  & Machine Translation & Avg.          & QA            & Summarization                     & Machine Translation & Avg.          \\ \hline
GPT-4o-0806                 & \textbf{86.6} & \textbf{75.5} & \textbf{85.6} & 86.4                & \textbf{84.8} & \textbf{89.6} & \textbf{91.2}                     & {\ul 92.3}          & \textbf{90.7} \\
GPT-4o-mini            & 82.9          & 58.9          & 82.3          & 79.6                & 78.9          & 84.2          & 86.4                              & 83.7                & 84.8          \\
Llama3.1-8B-Instruct  & 72.3          & 60.2          & 62.6          & 68.3                & 68.6          & -             & -                                 & -                   & -             \\
Llama3.1-70B-Instruct & 83.2          & 75.2          & 80.9          & 73.3                & 80.3          & -             & -                                 & -                   & -             \\
Llama3-8B-chinese     & -             & -             & -             & -                   & -             & 57.0          & 71.8                              & 55.5                & 61.6          \\
Llama3-70B-chinese    & -             & -             & -             & -                   & -             & 75.4          & 82.3                              & 68.8                & 76.2          \\
Qwen2-7B-Instruct     & 64.2          & 56.8          & 66.4          & 74.8                & 64.9          & 66.4          & 74.1                              & 72.3                & 70.3          \\
Qwen2-72B-Instruct    & 82.7          & 73.6          & 77.0          & 82.1                & 80.5          & 82.0          & 88.5                              & 82.3                & 84.2          \\
Qwen2.5-7B-Instruct   & 71.6          & 66.1          & 72.8          & 80.9                & 72.3          & 78.4          & 79.9                              & 76.2                & 78.4          \\
Qwen2.5-14B-Instruct  & 79.8          & 73.1          & 79.6          & 87.2                & 79.8          & 84.5          & 89.0                              & 85.2                & 86.1          \\
Qwen2.5-72B-Instruct  & {\ul 85.7}    & {\ul 74.7}    & 78.7          & 86.6                & 83.3          & 86.7          & 90.8                              & 88.8                & 88.5          \\
Lynx-8B-v1.1          & 83.2          & 66.1          & 65.8          & 74.0                & 76.9          & -             & -                                 & -                   & -             \\ \hline
Bi'an-qwen-7B         & 80.7          & 66.0          & 82.4          & {\ul 90.3}          & 80.2          & 85.9          & 88.4                              & 90.3                & 87.7          \\
Bi'an-qwen-14B        & 84.5          & 69.6          & {\ul 83.7}    & \textbf{92.5}       & {\ul 83.4}    & {\ul 88.4}    &  {\ul 90.9} & \textbf{92.6}       & {\ul 90.1}    \\ \bottomrule
\end{tabular}
}
\caption{Performance comparison on Bi'anBench\_EN and Bi'anBench\_ZH. The native Llama3 series (including Lynx fine-tuned based on it) exhibits weak support for Chinese, resulting in missing results for Bi'anBench\_ZH. Therefore, we conduct supplementary experiments using the open-source, continued pre-training version of Llama3 in Chinese. Here, we present the average results on the corresponding subsets for each task, while \texttt{Avg.} represents the average result across all subsets (not tasks).  We run experiments three times and report the average score. The best results are in bold, and the second-best results are underlined.}
\label{tab:main_exp}
\end{table*}

We report the results of different models on Bi'anBench\_EN and Bi'anBench\_ZH, categorized by task in Table \ref{tab:main_exp}. For more detailed results at the subset level, please refer to the Appendix \ref{app:exp}.

Firstly, we observe a significant variation in model performance based on size, with RAG hallucination detection remaining a challenging task for small-scale models. Secondly, our targeted training approach has yielded substantial improvements in hallucination detection capabilities. Notably, the Bi'an-qwen-7B model demonstrates superior performance compared to GPT-4o-mini, while the enhanced Bi'an-qwen-14B not only outperforms Qwen2.5-72B-Instruct but also narrows the performance gap with the state-of-the-art GPT-4o model, positioning itself as a viable, cost-efficient solution for production deployment. However, the results of subset performance indicate that Bi'an models still lag behind GPT-4o in numerical computation and long-context processing, indicating room for further optimization.
\section{Analysis}
\subsection{Ablation Study}
\begin{figure}[htbp]
    \centering
    \includegraphics[width=1.0\linewidth]{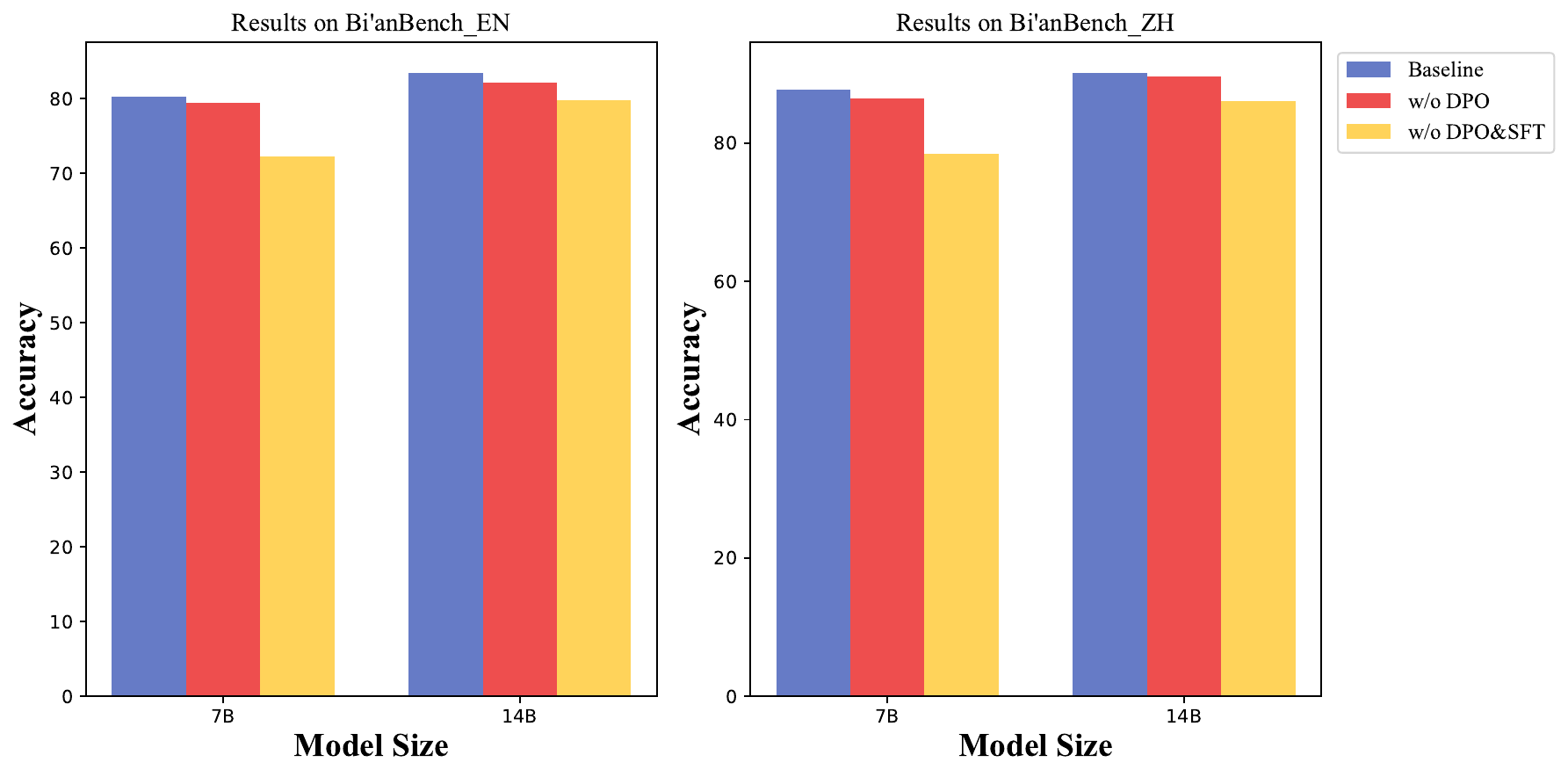}
    \caption{Ablation studies on the training phase of the Bi'an models.}
    \label{fig:ablation}
\end{figure}

As we employ a two-stage training process for Bi'an models, we conduct ablation studies on the training phase to validate its effectiveness. As shown in Figure \ref{fig:ablation}, our training method yields cumulative gains for both the 7B and 14B models, with the gains in the SFT stage being significantly higher than those in the DPO training stage.

\subsection{RQ: Impact of Knowledge Conflicts}

\begin{table}[]
\centering
\scalebox{0.7}{
\begin{tabular}{ccc}
\toprule
Model                & \multicolumn{2}{c}{Bi'anBench\_CF} \\ \hline
                     & EN               & ZH              \\ \hline
GPT-4o-0806          & 89.3             & 93.9            \\
GPT-4o-mini          & 92.8             & \textbf{95.7}   \\
Qwen2.5-7B-Instruct  & 93.2             & 87.3            \\
Qwen2.5-14B-Instruct & 91.8             & 80.8            \\
Qwen2.5-72B-Instruct & 91.7             & 94.6            \\ \hline
Bi'an-qwen-7B        & \textbf{94.1}    & {\ul 95.1}      \\
Bi'an-qwen-14B       & {\ul 93.3}       & 93.3            \\ \bottomrule
\end{tabular}
}
\caption{Performance comparison on Bi’anBench\_CF.}
\label{tab:cf}
\end{table}

Although research exists on the impact of conflicts between parametric knowledge and contextual knowledge on RAG \citep{longpre-etal-2021-entity, 10.5555/3692070.3694543}, the influence of this conflict on RAG hallucination detection remains under-researched. We conduct experiments on the constructed counterfactual QA hallucination detection dataset, Bi'anBench\_CF, and the results are presented in Table \ref{tab:cf}. An intriguing finding emerged: smaller models may outperform larger ones in certain scenarios, suggesting that the richer parametric knowledge of larger models could interfere with hallucination detection. We further manually annotate the bad cases of GPT-4o, revealing that 43.9\% of errors are directly related to parametric knowledge; detailed findings are provided in Appendix \ref{sec:cf}. Future research should explore how to mitigate the impact of such conflicts.
\section{Conclusion}
In this paper, we introduce Bi'an, a framework featuring a bilingual benchmark and lightweight judge models for RAG hallucination detection.

\clearpage
\section{Limitations}
\textbf{Loss of samples.} In the construction of training samples, the sample is discarded if all models predict incorrectly, leading to a loss of samples, particularly hard ones. Although we have reduced the overall loss rate through the ensemble-based sample construction strategy (refer to Appendix \ref{sec:tdc}), some loss is inevitable. We provide further discussion in Appendix \ref{sec:back}.


\noindent \textbf{Task Coverage.} As mentioned in Section \ref{sec:bench}, we have excluded creative writing from the scope of tasks due to its inconsistency with our operational definition of RAG hallucinations. However, from a practical application standpoint, creative writing with certain analytical or subjective content is a significant scenario for RAG. Detecting hallucinations in creative writing requires us to expand our evaluation framework further.

\bibliography{acl_latex}

\appendix
\section{Construction Details of Bi'anBench}
\label{bench_detail}
\subsection{Data Collection}
For the English subset of Bi'anBench, the question-answering section supplements the six source datasets (HaluEval \citep{li-etal-2023-halueval}, RAGTruth \citep{niu-etal-2024-ragtruth}, FinanceBench \citep{islam2023financebenchnewbenchmarkfinancial}, DROP \citep{dua-etal-2019-drop}, CovidQA \citep{moller-etal-2020-covid}, PubMedQA \citep{jin2019pubmedqadatasetbiomedicalresearch}) included in HaluBench with ASQA \citep{stelmakh-etal-2022-asqa} and IfQA \citep{yu-etal-2023-ifqa}. ASQA is a long-form factoid question-answering dataset, while IfQA is a dataset that requires hypothetical reasoning based on the information provided in the questions, thereby addressing the coverage limitations of HaluBench. The summarization section is derived from FIB \citep{tam-etal-2023-evaluating} and HaluEval's summarization subset. The Data-to-Text section comes from WebNLG \citep{gardent-etal-2017-creating} and RAGTruth's Data-to-Text subset. The machine translation section is sourced from PDC \citep{sun-etal-2022-rethinking} and WMT21 \citep{akhbardeh-etal-2021-findings}.

For the Chinese subset of Bi'anBench, the question-answering section is derived from a subset of document-based QA from CRUD \citep{10.1145/3701228}, WebQA1.0 \citep{li2016datasetneuralrecurrentsequence}, the reading comprehension section of LawBench \citep{fei-etal-2024-lawbench}, and WebCPM-LFQA \citep{qin-etal-2023-webcpm}. The summarization section is sourced from CSDS \citep{lin-etal-2021-csds}, the summarization subset of CRUD, and the summarization subset of LawBench. The machine translation section is drawn from PDC and WMT21.

For the counterfactual QA subset of Bi'anBench, the English portion is sourced from ConflictQA \citep{xie2024adaptive}, while the Chinese portion is derived from WebQA1.0.

The details of the data composition for the above three subsets are listed in Tables \ref{tab:bian_en}, \ref{tab:bian_zh}, and \ref{tab:bian_cf}, respectively.

\begin{table*}[]
\scalebox{0.89}{
\begin{tabular}{ccccc}
\hline
Dataset       & Domain   & Task                & Sampling size & Need perturbation? \\ \hline
HaluEval\_qa \citep{li-etal-2023-halueval} & General  & QA                  & 1,000         & No                 \\
RAGTruth\_qa \citep{niu-etal-2024-ragtruth} & General  & QA                  & 893           & No                 \\
FinanceBench \citep{islam2023financebenchnewbenchmarkfinancial} & Finance  & QA                  & 1,000         & No                 \\
DROP   \citep{dua-etal-2019-drop}       & Wikipedia     & QA                  & 995           & No                 \\
CovidQA   \citep{moller-etal-2020-covid}    & Medicine & QA                  & 970           & No                 \\
PubMedQA   \citep{jin2019pubmedqadatasetbiomedicalresearch}   & Medicine & QA                  & 1,000         & No                 \\
ASQA   \citep{stelmakh-etal-2022-asqa}       & Wikipedia     & QA                  & 1,000         & Yes                \\
IfQA   \citep{yu-etal-2023-ifqa}       & Wikipedia     & QA                  & 995           & Yes                \\
FIB    \citep{tam-etal-2023-evaluating}       & News     & Summarization       & 848           & No                 \\
HaluEval\_sum  \citep{li-etal-2023-halueval} & General  & Summarization       & 1,000         & No                 \\
WebNLG   \citep{gardent-etal-2017-creating}     & General  & Data-to-Text        & 1,100         & Yes                \\
RAGTruth\_d2t \citep{niu-etal-2024-ragtruth} & General  & Data-to-Text        & 500           & No                 \\
PDC     \citep{sun-etal-2022-rethinking}      & News     & Machine Translation & 1,000         & Yes                \\
WMT21   \citep{akhbardeh-etal-2021-findings}      & News     & Machine Translation & 1,000         & Yes                \\ \hline
\end{tabular}
}
\caption{Data composition for Bi'anBench\_EN.}
\label{tab:bian_en}
\end{table*}

\begin{table*}[]
\scalebox{0.85}{
\begin{tabular}{ccccc}
\hline
Dataset       & Domain     & Task                & Sampling size & Need perturbation? \\ \hline
CRUD\_MDQA \citep{10.1145/3701228}   & News       & QA                  & 937           & Yes                \\
WebQA 1.0  \citep{li2016datasetneuralrecurrentsequence}   & General    & QA                  & 977           & Yes                \\
LawBench\_RC \citep{fei-etal-2024-lawbench} & Law        & QA                  & 439           & Yes                \\
WebCPM-LFQA \citep{qin-etal-2023-webcpm}  & General    & QA                  & 983           & Yes                \\
CSDS   \citep{lin-etal-2021-csds}       & E-commerce & Summarization       & 996           & Yes                \\
CRUD\_sum  \citep{10.1145/3701228}   & News       & Summarization       & 949           & Yes                \\
LawBench\_sum \citep{fei-etal-2024-lawbench} & Law        & Summarization       & 476           & Yes                \\
PDC   \citep{sun-etal-2022-rethinking}        & News       & Machine Translation & 1,000         & Yes                \\
WMT21    \citep{akhbardeh-etal-2021-findings}      & News       & Machine Translation & 1,000         & Yes                \\ \hline
\end{tabular}
}
\caption{Data composition for Bi'anBench\_ZH.}
\label{tab:bian_zh}
\end{table*}

\begin{table*}[]
\scalebox{0.95}{
\begin{tabular}{ccccc}
\hline
Dataset    & Domain  & Task              & Sampling size & Need perturbation? \\ \hline
ConflictQA \citep{xie2024adaptive} & Wiki    & Counterfactual QA & 1000          & No                 \\
WebQA 1.0 \citep{li2016datasetneuralrecurrentsequence} & General & QA                & 934           & Yes                \\ \hline
\end{tabular}
}
\caption{Data composition for Bi'anBench\_CF.}
\label{tab:bian_cf}
\end{table*}

\subsection{Prompt Templates of Synthetic Data Generation Pipelines}
Prompt templates for the hallucination perturbation pipeline and counterfactual QA generation pipeline are shown in Table \ref{tab:prompt4perturb} and \ref{tab:prompt4cf}. We use the following prompts with GPT-4o-0806 with temperature=0.8.

We apply the hallucination perturbation pipeline to the data in Bi'anBench\_EN and Bi'anBench\_ZH that required perturbation. Since ConflictQA already meets the requirements, we only apply the counterfactual QA generation pipeline to WebQA1.0 when constructing Bi'anBench\_CF.

\begin{CJK*}{UTF8}{gbsn}
\begin{table*}[]
\centering
\small
\scalebox{1.0}{
\begin{tabular}{@{}ccp{10cm}@{}}
\toprule
\textbf{Task}                      & \textbf{Language}            & \textbf{Prompt}                                                                                                                                                                                                                                                                                                                                                                                                                                                                                                                                                                                                                                       \\ \midrule
\multirow{10}{*}{Question Answering}   & \multirow{5}{*}{EN}               & QUESTION:\textbackslash{}n \{\}\textbackslash{}n GOLD\_ANSWER:\textbackslash {}n \{\} \textbackslash{}n EVIDENCE\_TEXT:\textbackslash{}n \{\} \textbackslash{}n How can we change the GOLD\_ANSWER subtly such that it would be wrong? The perturbed answer should still give the impression of a valid answer, but inspection of the EVIDENCE\_TEXT would reveal that the perturbed answer is factually wrong.\textbackslash{}n Output the new answer and change made in JSON format with the key `new\_answer' and `change\_made'.                                                                                                                                                                                                                                                             \\
                            & \multirow{5}{*}{ZH}                & 问题: \textbackslash {}n \{\} \textbackslash{}n 标准答案:\textbackslash {}n \{\} \textbackslash{}n 证据文本:\textbackslash {}n \{\} \textbackslash{}n我们怎样才能巧妙地改变标准答案，这样它就会是错误的，且有足够的误导性？扰动后的答案仍应给出为有效答案的印象，但检查证据文本后会发现扰动的答案实际上是错误的，最多允许改动两处。\textbackslash{}n 请以JSON格式输出新答案和具体所做的更改，键为 `新答案' 和 `更改操作'。  \\
\hline
\multirow{5}{*}{Summarization} & \multirow{5}{*}{ZH}     & 标准摘要:\textbackslash {}n \{\} \textbackslash{}n 原始文本:\textbackslash {}n \{\} \textbackslash{}n 我们怎样才能巧妙地改变标准摘要，这样它就会是错误的，且有足够的误导性？扰动后的摘要仍应给出为有效摘要的印象，但检查原始文本后会发现扰动的摘要实际上是错误的，最多允许改动两处。\textbackslash{}n 请以JSON格式输出新摘要和具体所做的更改，键为 `新摘要' 和 `更改操作'。                                                                                                                                                                                                                                                                                                                                                                                                 \\
\hline
\multirow{4}{*}{Data-to-Text}                   & \multirow{4}{*}{EN} & TARGET\_TEXT:\textbackslash {}n \{\} \textbackslash{}n RAW\_DATA:\textbackslash {}n \{\} \textbackslash{}n How can we change the TARGET\_TEXT subtly such that it would be wrong? The perturbed text should still give the impression of a valid description, but inspection of the RAW\_DATA would reveal that the perturbed text is factually wrong.\textbackslash{}n Output the new text and change made in JSON format with the key `new\_text' and `change\_made'.     \\

\hline
\multirow{10}{*}{Machine Translation} & \multirow{5}{*}{EN}     & SOURCE\_TEXT:\textbackslash {}n \{\} \textbackslash{}n TRANSLATED\_TEXT:\textbackslash {}n \{\} \textbackslash{}n How can we change the TRANSLATED\_TEXT subtly such that it would be wrong? The perturbed text should still give the impression of a valid translation, but inspection of the SOURCE\_TEXT would reveal that the perturbed text is factually wrong.\textbackslash{}n You may incorporate seemingly relevant but actually nonexistent details, or alter the entity, figures, and time as needed; however, modifications are limited to no more than two instances. \textbackslash{}n Output the new text and change made in JSON format with the key `new\_text' and `change\_made'.                                                                                                                                                                     \\
                            & \multirow{5}{*}{ZH}       & 源文本:\textbackslash {}n \{\} \textbackslash{}n 译文:\textbackslash {}n \{\} \textbackslash{}n 我们怎样才能巧妙地改变译文，这样它就会是错误的，且有足够的误导性？扰动后的译文仍应给出为正确译文的印象，但检查源文本后会发现扰动的译文实际上是错误的。你可以使用增加似乎相关但实际上不存在的细节信息，或者改变主体、数值、时间等操作，但最多允许改动两处。\textbackslash{}n 请以JSON格式输出新译文和具体所做的更改，键为 `新译文' 和 `更改操作'。

\\ \bottomrule
\end{tabular}
}
\caption{Prompt templates for hallucination perturbation pipeline.}
\label{tab:prompt4perturb}
\end{table*}
\end{CJK*}

\begin{CJK*}{UTF8}{gbsn}
\begin{table*}[]
\centering
\small
\scalebox{1.0}{
\begin{tabular}{@{}ccp{10cm}@{}}
\toprule
\textbf{Phase}           & \textbf{Language}              & \textbf{Prompt}                                                                                                                                                                                                                                                                                                                                                                                                                                                                                                                                                                                                                                       \\ \midrule
\multirow{2}{*}{Perturb}      & \multirow{2}{*}{ZH}            & 对于当前问题“\{\}”，已知其正确答案为“\{\}”，请给出一个回答问题“\{\}”的极具迷惑性的错误答案，并给出能够得出错误答案的证据信息。你的输出应当是JSON格式的，键名为“错误答案”和“证据”。      \\                                                                                                                                                                                                                                                 
\hline
\multirow{4}{*}{Expand}  & \multirow{4}{*}{ZH}  & 你现在将要忠实地进行文本扩写任务。请将文本“\{\}”扩写为一段300字以上的段落，段落中必须忠实准确地包含了文本信息，且不得含有任何与文本矛盾的信息。你无需考虑文本本身的真实性而进行纠错。                                                                                                                                                                                                                                                                                                                                                                                                 \\
 \bottomrule
\end{tabular}
}
\caption{Prompt templates for counterfactual QA generation pipeline.}
\label{tab:prompt4cf}
\end{table*}
\end{CJK*}
\subsection{Statistics of Bi'anBench}
Bi'anBench consists of a total of 22,992 test cases, divided into three subsets: the English subset Bi'anBench\_EN (13,301), the Chinese subset Bi'anBench\_ZH (7,757), and a special counterfactual QA subset Bi'anBench\_CF (1,934). Statistical details are listed in Table \ref{tab:bench_stat}. We present some examples of Bi'anBench in Table \ref{tab:bench_example}.

\begin{table*}[htbp]
\centering
\scalebox{1.0}{
\begin{tabular}{ccccccc}
\toprule
Subset         & Size   & \#PASS & \#FAIL & Avg\_len & Max\_len & Min\_len \\ \hline
Bi'anBench\_EN & 13,301 & 6,777  & 6,524  & 752.3    & 6,732    & 20       \\
Bi'anBench\_ZH & 7,757  & 3,757  & 4,000  & 1,103.3  & 32,696   & 37       \\
Bi'anBench\_CF & 1,934  & 952    & 982    & 246.6    & 504      & 38       \\
Overall        & 22,992 & 11,486 & 11,506 & 828.2    & 32,696   & 20       \\ \bottomrule
\end{tabular}
}
\caption{Statistics of Bi'anBench. \texttt{\#PASS} refers to the number of cases marked as hallucination-free in each data subset, while \texttt{\#FAIL} refers to the number of cases marked as containing hallucinations in each data subset. The length of a case is defined as the length of the token array obtained by concatenating the \textbf{context} and \textbf{answer} as strings and then tokenizing them using the Qwen2.5-14B-Instruct tokenizer.}
\label{tab:bench_stat}
\end{table*}

\begin{CJK*}{UTF8}{gbsn}
\begin{table*}[]
\centering
\small
\scalebox{1.0}{
\begin{tabular}{@{}ccp{10cm}@{}}
\toprule
\textbf{Task}           & \textbf{Language}              & \textbf{Example}                                                       \\ \midrule
\multirow{30}{*}{Question Answering}      & \multirow{10}{*}{EN}            & \{"id": 13, "task": "Question Answering", "context": "As of the census of 2000, there were 218,590 people, 79,667 households, and 60,387 families residing in the county.  The population density was 496 people per square mile (192/km²). There were 83,146 housing units at an average density of 189 per square mile (73/km²). The racial makeup of the county was 86.77\% Race (United States Census), 9.27\% Race (United States Census), 0.23\% Race (United States Census), 1.52\% Race (United States Census), 0.06\% Race (United States Census), 0.69\% from Race (United States Census), and 1.47\% from two or more races.  1.91\% of the population were Race (United States Census) or Race (United States Census) of any race. 22.5\% were of German people, 13.1\% Irish people, 9.8\% Italian people, 9.2\% English, 8.1\% \"American\" and 6.0\% Polish ancestry.", "question": "Which group from the census is smaller: German or Irish?", "answer": "German", "label": "FAIL", "source\_ds": "DROP"\}      \\                                                                                                       & \multirow{15}{*}{ZH}            & \{"id": 213, "task": "Question Answering", "context": ["新闻【1】：2023-08-15 20:22，正文：新华社洛杉矶8月14日电（记者谭晶晶 高山）美国夏威夷州州长乔希·格林14日表示，夏威夷州毛伊县野火遇难人数已升至99人。随着搜救工作的继续展开，死亡人数很可能继续上升。毛伊县警察局局长约翰·佩尔蒂埃说，目前99名遇难者中只有三具遗体通过指纹识别了身份。由于遗体在野火中损毁严重，遇难者身份鉴别工作难度极大。格林表示，此次野火已导致毛伊县超过2200栋建筑被损毁，其中约86\%是住宅楼，造成损失预计近60亿美元，这很可能是该州历史上最惨重的一次自然灾难。 美国夏威夷州毛伊县8月10日公布的照片显示野火造成严重破坏。新华社发根据毛伊县政府当日发布的最新通报，代号为“普莱胡”的野火已被扑灭；代号为“拉海纳”的野火已有85\%得到控制，目前“没有活跃的威胁”；代号为“毛伊内陆”的野火已有60\%得到控制。据通报介绍，野火火势100\%得到控制并不意味着其已被扑灭，仅表明消防员已将火势控制在一定范围内，但野火在此范围内仍可能继续燃烧。当消防人员认为已没有任何物体继续燃烧时，才会宣布野火被扑灭。毛伊岛是夏威夷群岛第二大岛，岛上人口超过10万。据夏威夷紧急事务管理部门消息，野火始于8日，受飓风带来的强风以及环境干旱影响迅速蔓延。"], "question": "夏威夷州毛伊县发生的野火导致多少人遇难，以及多少栋建筑被损毁？", "answer": "夏威夷州毛伊县的野火导致95人遇难，超过2100栋建筑被损毁。", "label": "失败", "source\_ds": "CRUD\_MDQA"\}      \\                                                                                                                                          
\hline
\multirow{5}{*}{Data-to-Text}  & \multirow{5}{*}{EN}  & \{"id": 11071, "task": "Data-to-Text", "context": "['Alhambra | shipBeam | 8.3 m' 'Alhambra | builder | Samuda\_Brothers'\textbackslash{}n 'Alhambra | length | 63800.0 (millimetres)'\textbackslash{}n 'Alhambra | status | \"Wrecked\"' 'Alhambra | topSpeed | 18.52']", "question": "", "answer": "Before it wrecked, the Alhambra was built by the Samuda Brothers. It had a length of 63800.0 mms, a top speed of 18.52 and a ship beam of 8.3 metres.", "label": "PASS", "source\_ds": "WebNLG"\}                                                                                                                                                                                                                                                                                                                                                                                                 \\
\hline
\multirow{5}{*}{Machine Translation}  & \multirow{5}{*}{ZH}  & \{"id": 30004, "task": "Machine Translation", "context": "by now you should agree that multimedia isn \&apos;t any one thing but a complex entity that involves the many things : hardware , software , and the interface where they meet .", "question": "", "answer": "它不只是一件东西，而是包括许多东西的复杂的组合：硬件、软件和这两者相遇时的界面。", "label": "通过", "source\_ds": "wmt21", "type": "en2zh"\}                                                                                                                                                                                                                                                                                                                                                                                              \\
 \bottomrule
\end{tabular}
}
\caption{Examples of Bi'anBench. Each case is structured as a JSON object, comprising several key components: the \texttt{id} field uniquely identifies each data entry, \texttt{task} specifies the type of task being performed, and \texttt{context} supplies the relevant background text. The \texttt{question} field contains the query text, which remains empty except in question-answering (QA) tasks. The \texttt{answer} field holds the response text that requires hallucination detection, while \texttt{label} indicates the correct label (hallucination-free as `PASS' or `通过' while hallucination-included as `FAIL' or `失败'). Additionally, \texttt{source\_ds} denotes the origin dataset from which the case is derived. For machine translation tasks, an extra \texttt{type} field is included to specify the direction of translation (e.g., source-to-target language). }
\label{tab:bench_example}
\end{table*}
\end{CJK*}

\section{Model Development}
\label{app:model_dev}
\subsection{Training Dataset Construction}
\label{sec:tdc}
We conduct stratified sampling on the source datasets in Bi'anBench to construct the initial data, ensuring that the sampled results are not simultaneously present in Bi'anBench to prevent test data leakage. The composition of the initial sampled data is shown in Table \ref{tab:init_data}. It is noteworthy that since the complete dataset of FinanceBench is not publicly available and thus cannot be sampled, we used a similar numerical computation benchmark, TableBench \citep{wu2024tablebenchcomprehensivecomplexbenchmark}, as a substitute.

After obtaining the initial data, we first generate a portion of hallucination-containing instances using the hallucination perturbation pipeline, similar to the benchmark construction process. Subsequently, both hallucination-included and hallucination-free instances are constructed as RAG hallucination detection requests and input in parallel into three models (GPT-4o-0806, GPT-4o-mini, Qwen2.5-72B-Instruct). The prompt templates for the requests are shown in Table \ref{tab:prompt4test}. The output format is JSON containing the reasoning process and the judgment result, which we parse to compare the judgment result with the ground truth label.

When constructing SFT samples, if there is only one correct prediction, we use that JSON as the sample's output. If there are multiple correct predictions, we select the best one based on the priority order: GPT-4o-0806 > Qwen2.5-72B-Instruct > GPT-4o-mini. If there are no correct predictions, the sample is discarded. The statistical data from this process is presented in Table \ref{tab:ensemble}, which provides empirical evidence that our ensemble learning strategy effectively harnesses the complementary strengths of different models.

In constructing preference pairs, we treat the best output selected during SFT sample construction and the other incorrect outputs as preference pairs. If all predictions are either correct or incorrect, no preference pair is constructed.

Table \ref{tab:train_example} provides examples from the training set. We ultimately obtained 5,994 SFT samples and 1,713 preference pairs.

\begin{table}[]
\small
\scalebox{0.8}{
\begin{tabular}{cccc}
\toprule
Dataset       & Language & Task                & Sampling size \\ \hline
PubMedQA      & EN       & QA                  & 580           \\
DROP          & EN       & QA                  & 581           \\
CovidQA       & EN       & QA                  & 583           \\
TableBench    & EN       & QA                  & 483           \\
FIB           & EN       & Summarization       & 472           \\
RAGTruth\_d2t & EN       & Data-to-Text        & 394           \\
WebQA 1.0     & ZH       & QA                  & 586           \\
WebCPM-LFQA   & ZH       & QA                  & 588           \\
CSDS          & ZH       & Summarization       & 588           \\
PDC           & EN\&ZH   & Machine Translation & 1,394         \\
Overall       &          &                     & 6,249         \\ 
\bottomrule
\end{tabular}
}
\caption{The composition of the initial sampled data.}
\label{tab:init_data}
\end{table}

\begin{CJK*}{UTF8}{gbsn}
\begin{table*}[]
\centering
\small
\scalebox{1.0}{
\begin{tabular}{@{}ccp{10cm}@{}}
\toprule
\textbf{Task}                      & \textbf{Language}            & \textbf{Prompt}                                                                                                                                                                                                                                                                                                                                                                                                                                                                                                                                                                                                                                             \\ \midrule
\multirow{10}{*}{Question Answering}   & \multirow{5}{*}{EN}               & Given the following QUESTION, DOCUMENT and ANSWER you must analyze the provided answer and determine whether it is faithful to the contents of the DOCUMENT.\textbackslash{}n The ANSWER must not offer new information beyond the context provided in the DOCUMENT. The ANSWER also must not contradict information provided in the DOCUMENT. Output your final verdict by strictly following this format: "PASS" if the answer is faithful to the DOCUMENT and "FAIL" if the answer is not faithful to the DOCUMENT. Show your reasoning.\textbackslash{}n --\textbackslash{}n QUESTION:\{\}\textbackslash{}n --\textbackslash{}n DOCUMENT:\{\}\textbackslash{}n--\textbackslash{}nANSWER:\{\}\textbackslash{}n--\textbackslash{}n Your output should be in JSON FORMAT with the keys "REASONING" and "SCORE": {{"REASONING": <your reasoning as bullet points>, "SCORE": <your final score>}}                                                                                                                                                                                                                                                             \\
                            & \multirow{5}{*}{ZH}                & 给定下列“问题”、“文档”和“答案”，你必须分析所提供的“答案”，并确定它是否忠实于“文档”的内容。\textbackslash{}n “答案”不得提供超出“文档”中提供的上下文的新信息。“答案”也不得与“文档”中提供的信息相矛盾。严格按照以下格式输出你的最终判断：如果“答案”忠实于“文档”，则为"通过" ；如果“答案”不忠实于“文档”，则为 "失败"。\textbackslash{}n --\textbackslash{}n 问题:\{\} \textbackslash{}n --\textbackslash{}n 文档:\{\} \textbackslash{}n--\textbackslash{}n 答案:\{\}\textbackslash{}n --\textbackslash{}n你的输出应该是 JSON 格式的，键是"推理过程"和"判断"：{{"推理过程":<你如何推理得到判断结果的步骤点>, "判断":<你的最终判断>}} \\
\hline
\multirow{10}{*}{Summarization} & \multirow{5}{*}{EN}     & Given the following DOCUMENT and SUMMARY you must analyze the provided summary and determine whether it is faithful to the contents of the DOCUMENT.\textbackslash{}n The SUMMARY must not offer new information beyond the context provided in the DOCUMENT. The SUMMARY also must not contradict information provided in the DOCUMENT. Output your final verdict by strictly following this format: "PASS" if the summary is faithful to the DOCUMENT and "FAIL" if the summary is not faithful to the DOCUMENT. Show your reasoning.\textbackslash{}n --\textbackslash{}n DOCUMENT:\{\}\textbackslash{}n--\textbackslash{}nSUMMARY:\{\}\textbackslash{}n--\textbackslash{}n Your output should be in JSON FORMAT with the keys "REASONING" and "SCORE": {{"REASONING": <your reasoning as bullet points>, "SCORE": <your final score>}}                                                                                                                                                                      \\
                            & \multirow{5}{*}{ZH}       & 给定下列“文档”和“摘要”，你必须分析所提供的“摘要”，并确定它是否忠实于“文档”的内容。\textbackslash{}n “摘要”不得提供超出“文档”中提供的上下文的新信息。“摘要”也不得与“文档”中提供的信息相矛盾。严格按照以下格式输出你的最终判断：如果“摘要”忠实于“文档”，则为"通过" ；如果“摘要”不忠实于“文档”，则为 "失败"。\textbackslash{}n --\textbackslash{}n 文档:\{\} \textbackslash{}n--\textbackslash{}n 摘要:\{\}\textbackslash{}n --\textbackslash{}n你的输出应该是 JSON 格式的，键是"推理过程"和"判断"：{{"推理过程":<你如何推理得到判断结果的步骤点>, "判断":<你的最终判断>}}                                                                                                                                                                                                                                      \\
\hline
\multirow{4}{*}{Data-to-Text}                   & \multirow{4}{*}{EN} & Given the following RAW\_DATA and TARGET\_TEXT you must analyze the provided target text and determine whether it is faithful to the contents of the RAW\_DATA.\textbackslash{}n The TARGET\_TEXT must not offer new information beyond the context provided in the RAW\_DATA. The TARGET\_TEXT also must not contradict information provided in the RAW\_DATA. Output your final verdict by strictly following this format: "PASS" if the target text is faithful to the RAW\_DATA and "FAIL" if the target text is not faithful to the RAW\_DATA. Show your reasoning.\textbackslash{}n --\textbackslash{}n RAW\_DATA:\{\}\textbackslash{}n--\textbackslash{}n TARGET\_TEXT:\{\}\textbackslash{}n--\textbackslash{}n Your output should be in JSON FORMAT with the keys "REASONING" and "SCORE": {{"REASONING": <your reasoning as bullet points>, "SCORE": <your final score>}}     \\

\hline
\multirow{10}{*}{Machine Translation} & \multirow{5}{*}{EN}     & Given the following SOURCE\_TEXT and TRANSLATED\_TEXT you must analyze the provided TRANSLATED\_TEXT and determine whether it is faithful to the contents of the SOURCE\_TEXT.\textbackslash{}n The TRANSLATED\_TEXT must not offer new information beyond the context provided in the SOURCE\_TEXT. The TRANSLATED\_TEXT also must not contradict information provided in the SOURCE\_TEXT. Output your final verdict by strictly following this format: "PASS" if the TRANSLATED\_TEXT is faithful to the SOURCE\_TEXT and "FAIL" if the TRANSLATED\_TEXT is not faithful to the SOURCE\_TEXT. Show your reasoning.\textbackslash{}n --\textbackslash{}n SOURCE\_TEXT:\{\}\textbackslash{}n--\textbackslash{}n TRANSLATED\_TEXT:\{\}\textbackslash{}n--\textbackslash{}n Your output should be in JSON FORMAT with the keys "REASONING" and "SCORE": {{"REASONING": <your reasoning as bullet points>, "SCORE": <your final score>}}                                                                                                                                                                      \\
                            & \multirow{5}{*}{ZH}       & 给定下列“源文本”和“译文”，你必须分析所提供的“译文”，并确定它是否忠实于“源文本”的内容。\textbackslash{}n “译文”不得提供超出“源文本”中提供的上下文的新信息。“译文”也不得与“源文本”中提供的信息相矛盾。严格按照以下格式输出你的最终判断：如果“译文”忠实于“源文本”，则为"通过" ；如果“译文”不忠实于“源文本”，则为 "失败"。\textbackslash{}n --\textbackslash{}n 源文本:\{\} \textbackslash{}n--\textbackslash{}n 译文:\{\}\textbackslash{}n --\textbackslash{}n你的输出应该是 JSON 格式的，键是"推理过程"和"判断"：{{"推理过程":<你如何推理得到判断结果的步骤点>, "判断":<你的最终判断>}}

\\ \bottomrule
\end{tabular}
}
\caption{Prompt templates used for RAG hallucination detection.}
\label{tab:prompt4test}
\end{table*}

\begin{table}[]
\begin{tabular}{cc}
\toprule
Model                & \#Correct \\ \hline
GPT-4o-0806          & 5,517         \\
GPT-4o-mini          & 5,146         \\
Qwen2.5-72B-Instruct & 5,364         \\ \hline
Ensemble             & 5,994         \\ \bottomrule
\end{tabular}
\caption{Statistical data of ensemble-based SFT sample construction.}
\label{tab:ensemble}
\end{table}
\end{CJK*}

\begin{CJK*}{UTF8}{gbsn}
\begin{table*}[]
\centering
\tiny
\scalebox{1.0}{
\begin{tabular}{@{}ccp{10cm}@{}}
\toprule
\textbf{Stage}                      & \textbf{Language}            & \textbf{Prompt}                                                                                                                                                                                                                                                                                                                                                                                                                                                                                                                                                                                                                                             \\ \midrule
\multirow{30}{*}{SFT}   & \multirow{20}{*}{EN}               & \{\textbf{'input'}: 'Given the following QUESTION, DOCUMENT and ANSWER you must analyze the provided answer and determine whether it is faithful to the contents of the DOCUMENT.\textbackslash{}n The ANSWER must not offer new information beyond the context provided in the DOCUMENT. The ANSWER also must not contradict information provided in the DOCUMENT. Output your final verdict by strictly following this format: "PASS" if the answer is faithful to the DOCUMENT and "FAIL" if the answer is not faithful to the DOCUMENT. Show your reasoning.\textbackslash{}n --\textbackslash{}n QUESTION:\textbackslash{}n Which person traveled with more ships to Malacca, Diogo Lopes de Sequeira or Albuquerque?\textbackslash{}n --\textbackslash{}n DOCUMENT:In 1509, Diogo Lopes de Sequeira was sent to Malacca by the king of Portugal with four ships in order to establish contact with the Sultanate of Malacca. Initially, Sequeira was well received by the Sultan Mahmud Syah . Soon however, the Muslim community convinced Mahmud Syah that the Portuguese should be eliminated. Several men were captured and killed, but the ships escaped. Albuquerque first departed from India for Malacca in April 1511, with 1,200 men and 17 to 18 ships. Albuquerque\'s objective was to sever Islamic trade and Venetian trade on the same occasion.A first attack by the Portuguese failed on 25 July 1511. Albuquerque\'s captains spoke against another attempt, but he struck again, succeeding in capturing Malacca in August, despite strong resistance and the presence of artillery on the Malaccan side. In celebration, Tristão da Cunha was sent to Pope Leo X in Rome with rich presents including the elephant that the pope named Hanno.\textbackslash{}n--\textbackslash{}n ANSWER:Sequeira\textbackslash{}n--\textbackslash{}n Your output should be in JSON FORMAT with the keys "REASONING" and "SCORE": \{"REASONING": <your reasoning as bullet points>, "SCORE": <your final score>\}', \textbf{'output'}: '\{"REASONING": ["The DOCUMENT states that Diogo Lopes de Sequeira traveled to Malacca with four ships.", "The DOCUMENT also states that Albuquerque traveled to Malacca with 17 to 18 ships.", "The QUESTION asks which person traveled with more ships to Malacca.", "The ANSWER provided is \'Sequeira\'.", "According to the DOCUMENT, Albuquerque traveled with more ships than Sequeira."], "SCORE": "FAIL"\}'\}                                                                                                                                                                                                                                                             \\ 
                            &  \multirow{10}{*}{ZH}                & \{\textbf{'input'}: '给定下列“文档”和“摘要”，你必须分析所提供的“摘要”，并确定它是否忠实于“文档”的内容。\textbackslash{}n “摘要”不得提供超出“文档”中提供的上下文的新信息。“摘要”也不得与“文档”中提供的信息相矛盾。严格按照以下格式输出你的最终判断：如果“摘要”忠实于“文档”，则为"通过" ；如果摘要不忠实于文档，则为 "失败"。\textbackslash{}n --\textbackslash{}n 文档:客服:请问下您有遇到什么问题需要我帮助您处理或者解决的呢?\textbackslash{}n用户:我想问一下我的发票什么时候寄回来\textbackslash{}n客服:这个订单电子发票已经开具了\textbackslash{}n客服:PC端:我的京东—客户服务—我的发票—发票详情下载即可;APP端:我的—客户服务—发票服务—发票详情查看即可\textbackslash{}n用户:我需要纸质的专票\textbackslash{}n用户:你再给我发一下我的订单，我给你选一下\textbackslash{}n客服:[订单编号]是这个订单吗\textbackslash{}n用户:不是\textbackslash{}n用户:[订单编号]\textbackslash{}n用户:是这个\textbackslash{}n客服:[订单编号]发票的运单号，已经在配送中了\textbackslash{}n用户:我能查一下到哪了吗?\textbackslash{}n客服:[站点]您的订单在京东[站点]发货完成，准备送往京东[站点]\textbackslash{}n用户:嗯，好的\textbackslash{}n用户:明天能到吗?\textbackslash{}n客服:妹子也没法保证的\textbackslash{}n用户:嗯，好的，[姓名]\textbackslash{}n客服:亲亲，客气了，妹子应该做的\textbackslash{}n客服:亲爱滴，请问还有其他需要帮助的么?\textbackslash{}n客服:如果没有的话，还请您记得给[姓名]评价哟，发财小手点一点哦\textbackslash{}n \textbackslash{}n--\textbackslash{}n 摘要:用户询问自己的发票什么时间能够寄回来，并表示自己需要纸质的专票，并询问是否明天能够到达。。客服回答用户的发票运单已经在京东[站点]发货完成，准备送往京东[站点]，但无法给用户保证是明天到货。\textbackslash{}n --\textbackslash{}n你的输出应该是 JSON 格式的，键是"推理过程"和"判断"：\{"推理过程":<你如何推理得到判断结果的步骤点>, "判断":<你的最终判断>\}', \textbf{'output'}: '\{"推理过程": ["用户询问发票什么时候寄回来，文档中客服回答电子发票已经开具，并提供了下载方式。", "用户表示需要纸质的专票，并提供了订单编号。", "客服确认了订单编号，并告知发票的运单号已经在配送中。", "用户询问是否可以查询到达位置，客服回答订单在京东站点发货完成，准备送往另一个京东站点。", "用户询问是否明天能到，客服表示无法保证。", "摘要中提到的内容与文档中的对话一致，没有提供超出文档的新信息，也没有与文档信息相矛盾。"], "判断": "通过"\}'\} \\
\hline
\multirow{35}{*}{Preference learning} & \multirow{20}{*}{EN}     & \{\textbf{'input'}: 'Given the following QUESTION, DOCUMENT and ANSWER you must analyze the provided answer and determine whether it is faithful to the contents of the DOCUMENT.\textbackslash{}n The ANSWER must not offer new information beyond the context provided in the DOCUMENT. The ANSWER also must not contradict information provided in the DOCUMENT. Output your final verdict by strictly following this format: "PASS" if the answer is faithful to the DOCUMENT and "FAIL" if the answer is not faithful to the DOCUMENT. Show your reasoning.\textbackslash{}n --\textbackslash{}n QUESTION:\textbackslash{}n What happened second: Arrêt de Mérindol or devastation of villages?\textbackslash{}n --\textbackslash{}n DOCUMENT:In 1545, Polin was on his way to fight against the English in the area of Boulogne. While in Marseilles in 1545, Polin was involved as a leader in the massacre of the Protestant Waldensians . Outside the Piedmont the Waldenses joined the local Protestant churches in Bohemia, France and Germany. After they came out of clandestinity and reports were made of sedition on their part, the French king, Francis I issued on 1 January 1545 the "Arrêt de Mérindol", and armed a crusade against the Waldensians of Provence. The leaders in the 1545 massacres were Jean Maynier d\'Oppède, First President of the parlement of Provence, and Antoine Escalin des Aimars who was returning from the Italian Wars with 2,000 veterans, the Bandes de Piémont. Deaths ranged from hundreds to thousands, depending on the estimates, and several villages were devastated.\textbackslash{}n--\textbackslash{}n ANSWER:[\'devastation of villages\']\textbackslash{}n--\textbackslash{}n Your output should be in JSON FORMAT with the keys "REASONING" and "SCORE": \{"REASONING": <your reasoning as bullet points>, "SCORE": <your final score>\}', \textbf{'chosen'}: '\{"REASONING": ["The DOCUMENT states that the \'Arrêt de Mérindol\' was issued on 1 January 1545.", "The DOCUMENT then mentions that a crusade was armed against the Waldensians of Provence.", "Following this, the DOCUMENT describes the involvement of leaders in the 1545 massacres and the devastation of several villages.", "Therefore, the \'Arrêt de Mérindol\' happened first, followed by the devastation of villages."], "SCORE": "PASS"\}', \textbf{'rejected'}: '\{"REASONING": ["The DOCUMENT states that the \'Arrêt de Mérindol\' was issued on 1 January 1545, and it led to a crusade against the Waldensians, resulting in the devastation of several villages.", "The QUESTION asks for the sequence of events, specifically what happened second: the \'Arrêt de Mérindol\' or the devastation of villages.", "The ANSWER only mentions \'devastation of villages\' without indicating its position in the sequence relative to the \'Arrêt de Mérindol\'.", "Since the DOCUMENT implies that the \'Arrêt de Mérindol\' was issued before the devastation of villages occurred, the ANSWER is incomplete and does not provide a clear sequence.", "The ANSWER does not contradict the DOCUMENT but fails to address the specific order of events as requested in the QUESTION."], "SCORE": "FAIL"\}'\}                                                                                                                                                                      \\
                            & \multirow{20}{*}{ZH}       & \{\textbf{'input'}: '给定下列“文档”和“摘要”，你必须分析所提供的“摘要”，并确定它是否忠实于“文档”的内容。\textbackslash{}n “摘要”不得提供超出“文档”中提供的上下文的新信息。“摘要”也不得与“文档”中提供的信息相矛盾。严格按照以下格式输出你的最终判断：如果“摘要”忠实于“文档”，则为"通过" ；如果摘要不忠实于文档，则为 "失败"。\textbackslash{}n --\textbackslash{}n 文档:用户:就是现在看到的就已经是最终优惠了的吗?\textbackslash{}n客服:您可以这么理解呢亲亲因为优惠我们都是接不到通知的呢亲亲商品支持价保呢亲亲请您放心\textbackslash{}n用户:确定不会明天变价格吧。\textbackslash{}n用户:??\textbackslash{}n客服:如果变价可以补回您差价呢亲亲\textbackslash{}n用户:明天如果有活动，\textbackslash{}n用户:我就亏了哦，\textbackslash{}n用户:如果价格已经是定了\textbackslash{}n用户:那我就买了哈\textbackslash{}n客服:不亏呀亲亲这个请您放心呀亲亲价格变动都是可以补给您差价的呢亲亲\textbackslash{}n用户:同时三个有优惠吗\textbackslash{}n用户:我要买三个\textbackslash{}n客服:目前这是小妹见过最低的价格了呢亲亲\textbackslash{}n用户:问个问题\textbackslash{}n客服:您请讲呢亲亲\textbackslash{}n用户:同时三个，\textbackslash{}n用户:在手机软件或者电脑软件上显示的时候\textbackslash{}n用户:能同时看到三个这样的吗\textbackslash{}n客服:电脑网页版可以呢亲亲手机不可以呢亲亲\textbackslash{}n用户:手机不可以吗?\textbackslash{}n用户:两个呢?\textbackslash{}n客服:手机只可以同时看一个呢亲亲\textbackslash{}n用户:手机能否上下，上面显示一个，下面显示一个?\textbackslash{}n客服:看店宝可以看呢亲亲\textbackslash{}n用户:因为两个店面，要同时监控\textbackslash{}n用户:??\textbackslash{}n客服:那推荐您购买看店宝呢亲亲\textbackslash{}n用户:有没有链接\textbackslash{}n用户:看店宝可以同时看是吗\textbackslash{}n客服:\textbackslash{}n客服:是的呢亲亲\textbackslash{}n客服:看店宝是一个全方位的屏幕呢亲亲可以分成四个屏幕查看呢亲亲等于四个镜头画面呢亲亲\textbackslash{}n客服:您看还有其他什么可以帮您的吗?小妹一定尽心尽力为您处理的哟\textbackslash{}n用户:那这个云台版的\textbackslash{}n用户:电脑的就可以是吗?\textbackslash{}n客服:云台版只有一个屏幕呢亲亲就是可以旋转呢亲亲\textbackslash{}n用户:可以同时看到几个屏幕\textbackslash{}n用户:电脑的\textbackslash{}n客服:云台版只有一个呀亲亲只是可以旋转呢亲亲\textbackslash{}n用户:刚才不是说电脑网页版可以吗?\textbackslash{}n客服:是的呢亲亲电脑网页版云台版最多可以同时查看一个呢亲亲\textbackslash{}n客服:手机只可以同时查看一个呢亲亲\textbackslash{}n客服:电脑网页版云台版最多可以同时查看六个呢亲亲\textbackslash{}n用户:哦哦，电脑网页的话就可以显示多个对吧\textbackslash{}n客服:对呢亲亲\textbackslash{}n \textbackslash{}n--\textbackslash{}n 摘要:用户询问客服商品后续是否会降价。客服请用户放心购买，如后续降价可参与价保。用户询问同时购买三个手是否可以在手机或电脑端同时看到三个。客服回答手机端可以，电脑端不可以，并建议用户购买看店宝。用户询问云台版的相关功能。客服为用户介绍云台版的相关功能，称其可以同时查看多个屏幕。\textbackslash{}n --\textbackslash{}n你的输出应该是 JSON 格式的，键是"推理过程"和"判断"：\{"推理过程":<你如何推理得到判断结果的步骤点>, "判断":<你的最终判断>\}', \textbf{'chosen'}: '\{"推理过程": ["摘要提到用户询问商品后续是否会降价，文档中确实有相关对话。", "摘要提到客服请用户放心购买，如后续降价可参与价保，文档中也有相关对话。", "摘要提到用户询问同时购买三个是否可以在手机或电脑端同时看到三个，文档中确实有相关对话。", "摘要提到客服回答手机端可以，电脑端不可以，文档中客服实际回答是手机端不可以，电脑端可以。", "摘要提到客服建议用户购买看店宝，文档中确实有相关对话。", "摘要提到用户询问云台版的相关功能，文档中确实有相关对话。", "摘要提到客服称云台版可以同时查看多个屏幕，文档中客服实际回答是云台版在电脑网页版可以同时查看六个屏幕。"], "判断": "失败"\}', \textbf{'rejected'}: '\{"推理过程": "摘要准确地反映了用户与客服之间的对话内容。用户询问了商品的降价情况，客服确认了价保政策。用户还询问了同时购买三个商品在手机和电脑上的显示情况，客服的回答与文档一致。最后，用户询问了云台版的功能，客服的回答也与文档中的信息相符。没有出现与文档内容相矛盾或超出文档范围的新信息，因此摘要忠实于文档。", "判断": "通过"\}'\}                                                                                                                                                                                        \\

\bottomrule
\end{tabular}
}
\caption{Examples of the training set.}
\label{tab:train_example}
\end{table*}
\end{CJK*}

\subsection{Hyperparameters of Training}
We have listed the key hyperparameters used during the training process in Table \ref{tab:hyper}. We utilize LLaMA-Factory \citep{zheng-etal-2024-llamafactory} as training framework. 

\begin{table}[]
\begin{tabular}{ccc}
\toprule
Hyperparameter                & SFT    & DPO    \\ \hline
batch\_size                   & 4      & 4      \\
gradient\_accumulation\_steps & 2      & 1      \\
learning\_rate                & 5.0e-5 & 5.0e-6 \\
epochs                        & 3.0    & 3.0    \\
lr\_scheduler\_type           & cosine & cosine \\
warmup\_ratio                 & 0.1    & 0.1    \\
lora\_rank                    & 8      & 8      \\
lora\_alpha                   & 16     & 16     \\ \bottomrule
\end{tabular}
\caption{Hyperparameters of Training.}
\label{tab:hyper}
\end{table}

\section{Experiments}
\label{app:exp}
The experimental setup employs identical prompt templates to those utilized in the training set construction, as detailed in Table \ref{tab:prompt4test}. For closed-source models, we access their official APIs, while for open-source models, we implement a local deployment configuration utilizing four Nvidia A100-80G GPUs, with vLLM-0.6.0 serving \citep{kwon2023efficient} as the inference engine. Throughout the inference process, we maintain consistent hyperparameters with top\_k=0.7 and temperature=0.8. To ensure statistical reliability, we conduct three independent runs and subsequently calculate the mean values for reporting purposes.
\subsection{Baselines}
We provide the source information for all tested baselines in Table \ref{tab:source}, with API addresses for closed-source models and HuggingFace repositories for open-source models.

\begin{table*}[]
\centering
\scalebox{0.8}{
\begin{tabular}{cc}
\toprule
Model                 & Source link                                                              \\ \hline
GPT-4o-0806           & \url{https://platform.openai.com/docs/models\#gpt-4o}                          \\
GPT-4o-mini           & \url{https://platform.openai.com/docs/models\#gpt-4o-mini }                    \\
Llama3.1-8B-Instruct  & \url{https://huggingface.co/meta-llama/Llama-3.1-8B-Instruct}                  \\
Llama3.1-70B-Instruct & \url{https://huggingface.co/meta-llama/Llama-3.1-70B-Instruct}                 \\
Llama3-8B-chinese     & \url{https://huggingface.co/shenzhi-wang/Llama3-8B-Chinese-Chat}               \\
Llama3-70B-chinese    & \url{https://huggingface.co/shenzhi-wang/Llama3-70B-Chinese-Chat}              \\
Qwen2-7B-Instruct     & \url{https://huggingface.co/Qwen/Qwen2-7B-Instruct}                            \\
Qwen2-72B-Instruct    & \url{https://huggingface.co/Qwen/Qwen2-72B-Instruct}                           \\
Qwen2.5-7B-Instruct   & \url{https://huggingface.co/Qwen/Qwen2.5-7B-Instruct}                          \\
Qwen2.5-14B-Instruct  & \url{https://huggingface.co/Qwen/Qwen2.5-14B-Instruct}                         \\
Qwen2.5-72B-Instruct  & \url{https://huggingface.co/Qwen/Qwen2.5-72B-Instruct}                         \\
Lynx-8B-v1.1          & \url{https://huggingface.co/PatronusAI/Llama-3-Patronus-Lynx-8B-Instruct-v1.1} \\ \bottomrule
\end{tabular}
}
\caption{Source links of all baselines.}
\label{tab:source}
\end{table*}

\subsection{Main Results}
We present the detailed experimental results at the subset level for Bi'anBench\_EN and Bi'anBench\_ZH in Tables \ref{tab:bianbench_en_exp} and \ref{tab:bianbench_zh_detail}.
\begin{table*}[]
\centering
\small
\scalebox{0.54}{
\begin{tabular}{cccccccccccccccc}
\toprule
Model                 & \multicolumn{8}{c}{Question Answering}                                               & \multicolumn{2}{c}{Summarization}  & \multicolumn{2}{c}{Data-to-Text} & \multicolumn{2}{c}{Machine Translation} & Average \\ \cline{2-15}
                      & HaluEval\_qa & RAGTruth\_qa & FinanceBench & DROP & CovidQA & PubMedQA & ASQA & IfQA & FIB  & HaluEval\_sum               & WebNLG      & RAGTruth\_d2t      & PDC                & WMT21              &         \\ \hline
GPT-4o-0806           & 86.9         & 84.0         & 85.7         & 83.1 & 95.0    & 83.9     & 87.1 & 86.7 & 74.4 & 76.6                        & 91.1        & 80.0               & 80.9               & 91.8               & 84.8    \\

GPT-4o-mini           & 87.2         & 80.2         & 80.7         & 79.8 & 86.5    & 84.2     & 80.2 & 84.3 & 60.1 & 57.6                        & 86.6        & 78.0               & 70.2               & 88.9               & 78.9    \\
Llama3.1-8B-Instruct  & 82.6         & 70.1         & 61.6         & 56.7 & 79.3    & 79.4     & 73.4 & 75.3 & 60.4 & 60.0                        & 84.5        & 40.6               & 57.0               & 79.6               & 68.6    \\
Llama3.1-70B-Instruct & 86.0         & 85.7         & 75.1         & 78.1 & 89.1    & 87.5     & 77.7 & 86.2 & 72.8 & 77.6                        & 90.5        & 71.2               & 59.2               & 87.3               & 80.3    \\
Qwen2-7B-Instruct     & 86.4         & 58.9         & 59.5         & 55.6 & 64.4    & 64.4     & 55.6 & 69.1 & 52.4 & 61.2                        & 78.8        & 54.0               & 66.2               & 83.3               & 64.9    \\
Qwen2-72B-Instruct    & 86.6         & 83.2         & 74.2         & 69.6 & 87.7    & 88.9     & 88.1 & 83.6 & 71.5 & 75.6                        & 91.8        & 62.2               & 74.8               & 89.3               & 80.5    \\
Lynx-8B-v1.1          & 86.2         & 79.4         & 76.5         & 72.7 & 96.3    & 88.4     & 82.0 & 83.7 & 62.1 & 70.0                        & 83.7        & 47.8               & 65.6               & 82.4               & 76.9    \\
Qwen2.5-7B-Instruct   & 85.3         & 74.6         & 59.3         & 54.0 & 82.5    & 75.3     & 64.9 & 76.8 & 64.4 & 67.8                        & 85.4        & 60.2               & 74.1               & 87.6               & 72.3    \\
Qwen2.5-14B-Instruct  & 83.4         & 78.4         & 73.2         & 71.2 & 90.7    & 81.8     & 78.7 & 81.1 & 70.4 & 75.8                        & 88.1        & 71.0               & 83.3               & 91.0               & 79.8    \\
Qwen2.5-72B-Instruct  & 86.2         & 83.7         & 78.4         & 81.3 & 92.8    & 88.9     & 88.5 & 85.8 & 72.3 & 77.0                        & 91.0        & 66.4               & 81.8               & 91.4               & 83.3    \\ \hline
Bi'an-qwen-7B         & 82.3         & 76.8         & 75.1         & 71.7 & 93.4    & 85.9     & 79.5 & 81.2 & 70.6 & 61.4                        & 87.4        & 77.4               & 91.3               & 89.3               & 80.2    \\
Bi'an-qwen-14B        & 85.7         & 80.7         & 79.3         & 78.5 & 93.3    & 87.5     & 86.1 & 84.5 & 69.7 & 69.4 & 86.0        & 81.4               & 94.5               & 90.5               & 83.4    \\ \bottomrule
\end{tabular}
}
\caption{Detailed experimental results for Bi'anBench\_EN.}
\label{tab:bianbench_en_exp}
\end{table*}

\begin{table*}[]
\centering
\small
\scalebox{0.695}{
\begin{tabular}{ccccccccccc}
\toprule
Model                & \multicolumn{4}{c}{Question Answering}           & \multicolumn{3}{c}{Summarization}                 & \multicolumn{2}{c}{Machine Translation} & Average \\ \cline{2-10}
                     & CRUD\_MDQA & WebQA1.0 & LawBench\_RC & WebCPM-LFQA & CSDS & CRUD\_sum                    & LawBench\_sum & PDC                & WMT21              &         \\ \hline
GPT-4o-0806                & 91.0      & 90.3     & 95.2        & 81.9        & 85.9 & 95.3                        & 92.4         & 92.4               & 92.2               & 90.7    \\

GPT4o-mini           & 86.8      & 87.6     & 91.6        & 70.9        & 76.1 & 92.1                        & 91.0         & 77.0               & 90.4               & 84.8    \\
Llama3-8B-chinese    & 63.8      & 58.5     & 70.0        & 35.8        & 63.0 & 73.7                        & 78.8         & 44.8               & 66.1               & 61.6    \\
Llama3-70B-chinese   & 82.1      & 85.8     & 89.6        & 44.1        & 72.7 & 83.1                        & 91.2         & 55.4               & 82.2               & 76.2    \\
Qwen2-7B-Instruct    & 71.2      & 79.5     & 73.0        & 41.9        & 63.7 & 78.1                        & 80.6         & 63.5               & 81.1               & 70.3    \\
Qwen2-72B-Instruct   & 89.0      & 89.6     & 90.8        & 58.7        & 81.3 & 90.6                        & 93.6         & 75.0               & 89.5               & 84.2    \\
Qwen2.5-7B-Instruct  & 86.7      & 82.4     & 87.9        & 56.7        & 71.6 & 89.6                        & 78.4         & 67.0               & 85.4               & 78.4    \\
Qwen2.5-14B-Instruct & 89.9      & 85.5     & 90.2        & 72.2        & 84.0 & 93.6                        & 89.5         & 79.1               & 91.2               & 86.1    \\
Qwen2.5-72B-Instruct & 90.8      & 90.5     & 92.9        & 72.5        & 84.1 & 94.1                        & 94.3         & 85.8               & 91.7               & 88.5    \\ \hline
Bi'an-qwen-7B        & 90.2      & 87.7     & 92.7        & 72.8        & 83.8 & 96.4                        & 85.1         & 90.6               & 90.0               & 87.7    \\
Bi'an-qwen-14B       & 93.4      & 90.3     & 94.1        & 75.8        & 86.9 & 96.1 & 89.8         & 93.4               & 91.8               & 90.1    \\ \bottomrule
\end{tabular}
}
\caption{Detailed experimental results for Bi'anBench\_ZH.}

\label{tab:bianbench_zh_detail}
\end{table*}

\section{Impact of Knowledge Conflicts}
\label{sec:cf}
\begin{CJK*}{UTF8}{gbsn}
In Table \ref{tab:bad_case_cf}, we provide examples of bad cases for GPT-4o-0806 on the Bi'anBench\_CF to demonstrate the impact of parametric knowledge on RAG hallucination detection. In case 1, the context explicitly states that "the capital of Romania is Iași," but the model corrects it to "the capital of Romania is Bucharest" based on its parametric knowledge, which affects the final judgment. Similarly, in case 2, the information provided is "韩愈是宋朝人," but the model determines "韩愈是唐朝人" through the reference to "长庆." We manually annotate all 57 bad cases of GPT-4o-0806 and find that 25 cases are directly related to interference from parametric knowledge, with a ratio of 25/57 = 43.9\%. This indicates that conflicts between parametric knowledge and contextual knowledge do indeed have an impact on RAG hallucination detection.

Additionally, the results of the Bi'an models in Table \ref{tab:cf} show that, even without specific training for counterfactual QA hallucination detection, the Bi'an models still demonstrate a significant improvement over the baseline models. This validates the effectiveness of our training approach.
\end{CJK*}
\begin{CJK*}{UTF8}{gbsn}
\begin{table*}[]
\centering
\small
\scalebox{1.0}{
\begin{tabular}{@{}ccp{11.5cm}@{}}
\toprule
Language             & Key        & Content                                                                                                                                                                                                                                                                                                                                                                                                                                                                                                                                                                                                                                                                                                                                                                                                                                                                                                                                                                                                                                                                                                                                                                                                                                                                                                                                                                                                                                                                                                                                                                                                                                           \\ \hline
                     & \multirow{15}{*}{context}    &  Iași is one of Romania's most important cities, with a rich history and profound cultural heritage. Over the centuries, {\color[HTML]{DF2A3F}Iași has served as the capital of Romania}, giving it a significant role on the national historical stage. Today, Iași continues to hold its important position as an economic, cultural, and transportation hub in the country. Located in the picturesque natural setting of the Moldova region in northeastern Romania, Iași is graced by the Prut River, which flows quietly around the historic city like a verdant ribbon, adding unique charm and serene scenery to the city. Iași not only holds significant economic importance but is also a thriving cultural city. It boasts numerous museums, libraries, and universities, the most famous being the University of Iași, which attracts a large number of scholars and students. The architectural styles in Iași are diverse, ranging from ancient medieval buildings to modern skyscrapers, coexisting in the city to showcase the historical trajectory and cultural features of different periods. As a city with convenient transportation, Iași is a major hub in Romania, with well-developed rail, road, and air transport, making it a crucial gateway connecting Romania with neighboring countries and other regions of Europe. The residents of Iași are warm and hospitable, and the city's traditional festivals and vibrant folk activities highlight the rich folk culture of the Moldova region. In conclusion, Iași is not only a city worth exploring but also a vibrant and charming cultural and economic center. \\
                     & question   & What is the name of the capital of Romania?                                                                                                                                                                                                                                                                                                                                                                                                                                                                                                                                                                                                                                                                                                                                                                                                                                                                                                                                                                                                                                                                                                                                                                                                                                                                                                                                                                                                                                                                                                                                                                                                       \\
                     & answer     & Iași                                                                                                                                                                                                                                                                                                                                                                                                                                                                                                                                                                                                                                                                                                                                                                                                                                                                                                                                                                                                                                                                                                                                                                                                                                                                                                                                                                                                                                                                                                                                                                                                                                              \\
                     & label      & PASS                                                                                                                                                                                                                                                                                                                                                                                                                                                                                                                                                                                                                                                                                                                                                                                                                                                                                                                                                                                                                                                                                                                                                                                                                                                                                                                                                                                                                                                                                                                                                                                                                                              \\
EN & \multirow{3}{*}{pred\_info} & \{"REASONING": "The document mentions that Iași is one of Romania's important cities and that it once served as the capital of Romania in history. However, the document does not explicitly state that Iași is the current capital of Romania. The question asks about the current capital of Romania, and the answer provided is Iași. {\color[HTML]{3531FF}According to common knowledge, the current capital of Romania is Bucharest, not Iași.}", "SCORE": "FAIL"\}                                                                                                                                                                                                                                                                                                                                                                                                                                                                                                                                                                                                                                                                                                                                                                                                                                                                                                                                                                                                                                                                                                                                                                                                \\ \hline
                     & \multirow{15}{*}{context}    & {\color[HTML]{DF2A3F}宋朝时期，著名的文学家与思想家韩愈（768～824），字退之，被世人尊称为韩昌黎}。他出生在河阳，也就是现今的河南省孟州市，但他的籍贯郡望则是昌黎郡，也就是如今的河北省昌黎县。韩愈在历史上具有举足轻重的地位，他不仅在文学创作方面卓有成就，同时也是一位深富哲思的思想家。他的才华与见解在当时及后世都赢得了高度的评价和赞誉。\textbackslash{}n\textbackslash{}n韩愈的一生充满了学术与思想的洪流，但也难逃世俗生活中的种种困境。长庆四年，也即宋朝的公元824年，由于身体困乏和疾病缠身，他不得不向朝廷告假。此后的日子里，他一直休养在家，但他的病情却每况愈下。在这一年的十二月二日，韩愈最终因病逝世，终止了他那波澜壮阔而又多彩的一生，他去世时年仅五十七岁。\textbackslash{}n\textbackslash{}n韩愈的离世，使文学与思想界失去了一位耀眼的巨星。然而，他的思想与文学作品，却如同璀璨的明珠，一直闪耀在中国文化的浩瀚天空中，至今依然在影响着一代代的文人墨客和学者们。韩愈不仅是那个时代的文人代表，他所倡导的儒家思想、个性解放和革新精神，至今仍具有重要的历史和文化价值，激励着后人不断追求进步与真理。                                                                                                                                                                                                                                                                                                                                                                                                                                                                                                                                                                                                                                                                                                                                                                                                                                                                                                                                                                                                                                                                                                         \\
                     & question   & 韩愈是哪一个朝代的人                                                                                                                                                                                                                                                                                                                                                                                                                                                                                                                                                                                                                                                                                                                                                                                                                                                                                                                                                                                                                                                                                                                                                                                                                                                                                                                                                                                                                                                                                                                                                                                                                                        \\
                     & answer     & 宋                                                                                                                                                                                                                                                                                                                                                                                                                                                                                                                                                                                                                                                                                                                                                                                                                                                                                                                                                                                                                                                                                                                                                                                                                                                                                                                                                                                                                                                                                                                                                                                                                                                 \\
                     & label      & 通过                                                                                                                                                                                                                                                                                                                                                                                                                                                                                                                                                                                                                                                                                                                                                                                                                                                                                                                                                                                                                                                                                                                                                                                                                                                                                                                                                                                                                                                                                                                                                                                                                                                \\
\multirow{-5}{*}{ZH} & \multirow{2}{*}{pred\_info} & \{"推理过程": {[}"文档中提到韩愈的生卒年是768～824年。",   "文档中提到韩愈在长庆四年（公元824年）因病逝世。",    "文档中提到长庆四年是宋朝的公元824年。",    {\color[HTML]{3531FF}"根据历史知识，长庆年号属于唐朝，而非宋朝。"},    "因此，文档中的信息表明韩愈是唐朝的人，而不是宋朝的人。"  {]},  "判断": "失败"\}"\}                                                                                                                                                                                                                                                                                                                                                                                                                                                                                                                                                                                                                                                                                                                                                                                                                                                                                                                                                                                                                                                                                                                                                                                                                                                                                                                                                                                                                                                               \\ \bottomrule
\end{tabular}
}
\caption{Bad cases of GPT-4o-0806 on Bi'anBench\_CF.}
\label{tab:bad_case_cf}
\end{table*}
\end{CJK*}
\section{Backward Reasoning for Sample Construction}
\label{sec:back}
To address the issue of loss of samples, we designed a training sample construction method based on "backward reasoning." This approach involves using the true labels as hints to input into the model, requiring the model to provide the reasoning process leading to the labels. The prompt templates are shown in Table \ref{tab:prompt_back}.

We construct training sets from the initial sampled data using the ensemble-based sample construction method and the backward reasoning-based method, which are then utilized for SFT on Qwen2.5-7B-Instruct with the same hyperparameters (Table \ref{tab:hyper}). The experimental results are presented in Table \ref{tab:com_back} and \ref{tab:com_back_zh}. The results indicate that, although the backward reasoning-based sample construction method increases the number of samples, it does not exhibit a significant advantage.

\begin{CJK*}{UTF8}{gbsn}
\begin{table*}[]
\centering
\small
\begin{tabular}{@{}cp{11.5cm}@{}}
\toprule
Language & Prompt Template                                                                                                                                                                                                                                                                                                                                                                                                                                                                                                                                                                                                                                                                                                                                                                                                                                                                                                         \\ \hline
\multirow{5}{*}{EN}      & Given the following QUESTION, DOCUMENT and ANSWER, it is known that the ANSWER is \{\texttt{IF\_FAITHFUL}\} to the QUESTION. Please carefully analyze the information within the QUESTION, DOCUMENT and ANSWER, and provide a reasoning process for why it is \{\texttt{IF\_FAITHFUL}\}. Faithfulness means that the ANSWER must not provide new information that goes beyond the context provided in the DOCUMENT and must not contradict the information provided in the DOCUMENT. Note that the given judgment result is certainly correct, so your reasoning process should not contradict it.\textbackslash{}n --\textbackslash{}n QUESTION:\{\}\textbackslash{}n--\textbackslash{}n DOCUMENT:\{\}\textbackslash{}n--\textbackslash{}n ANSWER:\{\} \textbackslash{}n--\textbackslash{}n Your output should be in JSON FORMAT with the key "REASONING": \{\{"REASONING": \textless{}your reasoning as bullet points\textgreater{}\}\} \\
\multirow{5}{*}{ZH}       & 现在对于“问题”、“文档”和“答案”，已知答案\{\texttt{IF\_FAITHFUL}\}于文档，请你仔细分析问题、文档和答案中的信息，给出为何\{\texttt{IF\_FAITHFUL}\}的推理过程。忠实指“答案”不得提供超出“文档”中提供的上下文的新信息，也不得与“文档”中提供的信息相矛盾。注意已知的判断结果一定是正确的，因此你的推理过程不应与其矛盾。\textbackslash{}n -- \textbackslash{}n 问题: \{\} \textbackslash{}n -- \textbackslash{}n 文档:\{\} \textbackslash{}n -- \textbackslash{}n 答案: \{\} \textbackslash{}n -- \textbackslash{}n 你的输出应该是 JSON 格式的，键是"推理过程"：\{\{"推理过程":\textless{}你如何推理得到判断结果的步骤点\textgreater{}\}\}                                                                                                                                                                                                                                                                                                                                                                                                                                                                           \\ \bottomrule
\end{tabular}
\caption{Prompt templates of backward reasoning-based sample construction method. The possible values for \texttt{IF\_FAITHFUL}  are ['faithful', 'not faithful'] or ['忠实', '不忠实'].}
\label{tab:prompt_back}
\end{table*}
\end{CJK*}

\begin{table*}[]
\centering
\small
\scalebox{0.54}{
\begin{tabular}{cccccccccccccccc}
\toprule
Model              & \multicolumn{8}{c}{Question Answering}                                               & \multicolumn{2}{c}{Summarization} & \multicolumn{2}{c}{Data-to-Text} & \multicolumn{2}{c}{Machine Translation} & Average \\ \cline{2-15}
                   & HaluEval\_qa & RAGTruth\_qa & FinanceBench & DROP & CovidQA & PubMedQA & ASQA & IfQA & FIB         & HaluEval\_sum       & WebNLG      & RAGTruth\_d2t      & PDC                & WMT21              &         \\ \hline
Ensemble           & 83.0         & 79.4         & 73.5         & 70.3 & 92.4    & 84.2     & 81.5 & 82.1 & 70.1        & 67.7                & 88.1        & 70.0               & 80.4               & 88.2               & 79.4    \\
Backward reasoning & 84.8         & 78.1         & 75.8         & 66.9 & 90.5    & 85.1     & 76.1 & 80.0 & 71.2        & 67.0                & 85.7        & 76.2               & 92.2               & 88.6               & 79.9    \\ \bottomrule
\end{tabular}
}
\caption{Performace comparison between ensemble-based method and  backward reasoning-based method on Bi'anBench\_EN. We do SFT on Qwen2.5-7B-Instruct.}
\label{tab:com_back}
\end{table*}

\begin{table*}[]
\centering
\small
\scalebox{0.695}{
\begin{tabular}{ccccccccccc}
\toprule
Model              & \multicolumn{4}{c}{Question Answering}             & \multicolumn{3}{c}{Summarization} & \multicolumn{2}{c}{Machine Translation} & Average \\ \cline{2-10}
                   & CRUD\_MDQA & WebQA1.0 & LawBench\_RC & WebCPM-LFQA & CSDS  & CRUD\_sum & LawBench\_sum & PDC                & WMT21              &         \\ \hline
Ensemble           & 89.0       & 87.0     & 90.4         & 70.8        & 83.0  & 95.5      & 85.5          & 88.4               & 89.5               & 86.5    \\
Backward reasoning & 88.6       & 86.7     & 92.9         & 67.9        & 81.8  & 94.5      & 85.5          & 88.0               & 87.3               & 85.9    \\ \bottomrule
\end{tabular}
}
\caption{Performace comparison between ensemble-based method and  backward reasoning-based method on Bi'anBench\_ZH. We do SFT on Qwen2.5-7B-Instruct.}
\label{tab:com_back_zh}
\end{table*}
\end{document}